\documentclass[runningheads]{llncs}
\usepackage[T1]{fontenc}
\usepackage{graphicx}
\usepackage{hyperref}
\usepackage{xcolor}
\renewcommand\UrlFont{\color{blue}\rmfamily}
\hypersetup{colorlinks, linkcolor={blue}, citecolor={blue}, urlcolor={blue}}

\usepackage{amsmath}
\usepackage[misc,geometry]{ifsym} 

\usepackage{xspace}
\newcommand{\memento}{\textsc{Memento}\xspace}
\newcommand{\abbrv}{\emph{\underline{M}arvelous \underline{E}xperi\underline{MEN}t \underline{TO}ol}\xspace}

\usepackage{listings}

\usepackage{tikz}
\usetikzlibrary{positioning,fit,calc}
\usetikzlibrary{arrows.meta}
\definecolor{mem}{HTML}{0000FF}

\begin{document}

\title{\memento: Facilitating Effortless, Efficient, and Reliable ML Experiments}
\titlerunning{\memento}

\author{Zac Pullar-Strecker\inst{1}\and
    Xinglong Chang\inst{1}\and
    Liam Brydon\inst{1}\and
    Ioannis Ziogas\inst{1,2}\and
    Katharina Dost\inst{1}\and
    J\"org Wicker\inst{1}\textsuperscript{(\Letter)}
}
\authorrunning{Z. Pullar-Strecker et al.}

\institute{University of Auckland, Auckland, New Zealand\\
        \and University of Mississippi, Oxford, MS, USA\\
        \email{\{zpul156, xcha011, lbry121, izio995\}@aucklanduni.ac.nz}\\
        \email{\{katharina.dost, j.wicker\}@auckland.ac.nz}
}

\maketitle

\begin{abstract}
Running complex sets of machine learning experiments is challenging and time-consuming due to the lack of a unified framework.
This leaves researchers forced to spend time implementing necessary features such as parallelization, caching, and checkpointing themselves instead of focussing on their project.
To simplify the process, in this paper, we introduce \memento, a Python package that is designed to aid researchers and data scientists in the efficient management and execution of computationally intensive experiments. \memento has the capacity to streamline any experimental pipeline by providing a straightforward configuration matrix and the ability to concurrently run experiments across multiple threads.
A demonstration of \memento is available at: \href{https://wickerlab.org/publication/memento/}{wickerlab.org/publication/memento}.


\keywords{
    Experimental pipeline 
    \and Parallel computing 
    \and Reliable ML
}
\end{abstract}

\section{Introduction}

As machine learning (ML) and data science continue to shape decision-making across a wide range of domains \cite{Sarker2021}, researchers and practitioners are facing the ever-growing challenge of designing and executing complex experiments efficiently and reliably \cite{Zoeller2021}. 
This challenge is particularly acute in machine learning, where the traditional experimental pipeline involves a complex set of intermediate steps, such as data pre-processing, model selection, hyperparameter tuning, and model evaluation. 
Benchmarking multiple options for each of these steps easily results in an overwhelming number of individual experiments, long waiting times when run sequentially, and the need to adjust scripts or restart entire sets of experiments when errors occur.

Aiming to avoid these challenges in practice, automated machine learning (AutoML) tools can help choose the best combination of steps in the ML pipeline automatically \cite{feurer2015efficient,karmaker2021automl,komer2014hyperopt,thornton2013auto}, but they may not be flexible enough to integrate new algorithms or match multiple libraries as is necessary for research.
In Python, scikit-learn's pipeline functionality \cite{Pedregosa2011} allows users to conveniently choose and chain the steps involved in a single experiment but falls short in handling the setup of entire experiment sets.

To date, there exists no unified framework for the parallelization of experimental setups beyond a narrow range of machine learning applications \cite{idowu2022asset}. 
Most often, researchers have to manually implement workflow features, such as caching and checkpointing, which introduce an added layer of complexity. Moreover, remedial corrections to potential errors occurring partway through the execution of a series of long experiments require tedious debugging.

\begin{figure*}[t]
    \centering
    \resizebox{\columnwidth}{!}{\input{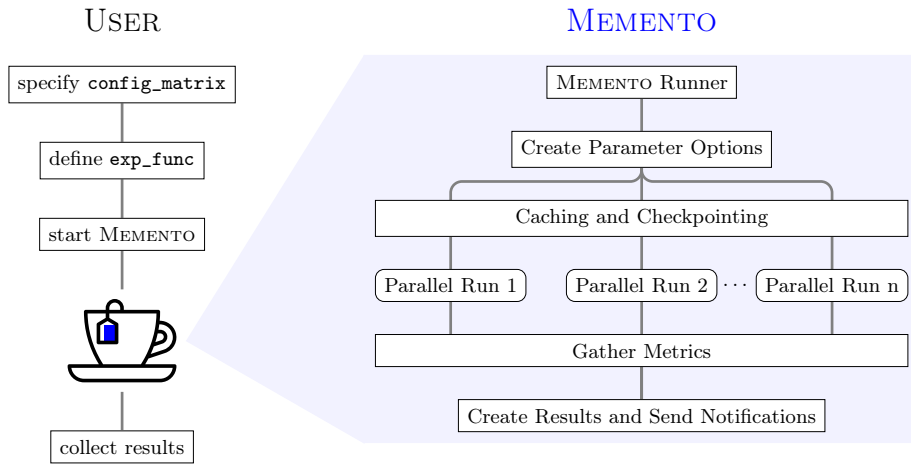}}
    \caption{The roles of the user and \memento when running a complex and large set of experiments with \memento.}
    \label{fig.workflow}
\end{figure*}

To fill this gap, we present \memento (\abbrv), a simple and flexible Python library designed to help researchers, data analysts, and machine learning practitioners streamline the experimental pipeline, parallelize experiments across threads, save and restore results, checkpoint in-progress experiments, and receive notifications when experiments fail or finish. 
By providing a unified framework for the parallelization of experimental setups, \memento can significantly reduce the amount of time and coding expertise required to structure and execute experimental workflows, as Figure \ref{fig.workflow} shows.

In the following sections, we present the main features of \memento, explain how it can help towards streamlining any experimental workflow, and conclude the paper, respectively.

\section{The \memento Package: Features and Strengths}

\memento is a modular, flexible, and easily configurable Python library that allows for the parallel execution of multiple experimental setups with the added benefits of automatic checkpoint creation, output caching, and error tracing. At the core of \memento's design lies a user-friendly and customizable configuration matrix outlining the building blocks of the intended experiments within a pipeline, while avoiding writing repetitive code. 
\memento automatically translates the matrix to distinct experimental tasks that can be allocated to different CPUs, thus effectively parallelizing the experimental pipeline and significantly reducing the time required for large-scale experiments. 

When running computationally intensive experiments, saving intermediate results and resuming the process from where it left off in case of unexpected failures or interruptions (e.g., power outages, hardware failures, or errors in a single task) is of utmost importance. \memento provides an easy way to do this through automated checkpointing. When checkpointing is enabled, \memento saves the experiment output at regular intervals, allowing for resumption without costly manual intervention, with the additional option of storing each result for post-evaluation.

The source code and documentation are available on GitHub\footnote{\memento source code and documentation: \href{https://github.com/wickerlab/memento}{\UrlFont{github.com/wickerlab/memento}}}, 
and the package can be installed via PyPI using \begin{verbatim}pip install memento-ml\end{verbatim}.

\section{Demonstration and Configuration}

To run a set of experiments with \memento, the user only needs to do the steps outlined in Figure \ref{fig.workflow} (left). After installing and importing \memento using
\begin{lstlisting}{}
    import memento
    # ...
\end{lstlisting}
the user needs to define a configuration matrix (\texttt{config\_matrix}). This matrix is the core of \memento and describes the list of experiments the user wants to run. One example is
\begin{lstlisting}
    # The configuration matrix conveniently specifies the experiments to be run.
    config_matrix = {
        "parameters": { 
            "dataset": [load_digits, load_wine, load_breast_cancer], 
            "feature_engineering": [DummyImputer, SimpleImputer],
            "preprocessing": [DummyPreprocessor, MinMaxScaler, StandardScaler],
            "model": [AdaBoost, RandomForest, SVC] 
        },
        "settings": {"n_fold": 5},
        "exclude": [{"dataset": load_digits, "feature_engineering": SimpleImputer}]
    }
\end{lstlisting}
\memento automatically constructs tasks using every combination of defined parameters, e.g., for all combinations of datasets, feature engineering methods, preprocessing techniques, and models above resulting in $3\times 2\times 3\times 3 = 54$ tasks.
The \texttt{settings} keyword stores constants that can be accessed by each task, removing the need to access global constants.
The keyword \texttt{exclude} is used as a lookup table to skip any unwanted combinations during the task generation allowing for a more fine-grained specification of required experiments.
The name and number of parameters can be fully customized, making \memento compatible with any type of machine-learning pipeline.
Each task is self-isolated and can be run in parallel.

Once the configuration matrix is defined, \memento passes every created task individually as a set of parameters into the user-defined experiment function (\texttt{exp\_func}), for example, 
\begin{lstlisting}
    # The experiment function is called for each individual set of parameters.
    def exp_func(context: memento.Context, config: memento.Config) -> any:
        if context.checkpoint_exist():  # Based on the hashing value.
            result = context.restore()  # Recover results from cache.
        else:
            # access the parameters for an individual experiment
            X, y = config.dataset() 
            model = config.model()
            n_fold = config.settings["n_fold"] 
            # prepare the experimental pipeline and run
            pipeline = make_pipeline(config.feature_engineering, config.preprocessing, model)
            result = cross_val_scores(pipeline, X, y, cv=n_fold)
            context.checkpoint(result)  # Caching results 
        return result
\end{lstlisting}
\memento is designed with fault tolerance and exception handling in mind. 
Each parameter is assigned a hash value when generating the tasks. 
If an error occurs during the execution, we can update the code and rerun it. 
To avoid running duplicate experiments, we specify to restore checkpoints if available. 
If not, we access the input parameters for this task specify the individual experiment that needs to be run. 
Lastly, we specify the outputs that should be checkpointed. 

Once both the configuration matrix and the experiment function are set up, the user can start \memento:
\begin{lstlisting}
    # start MEMENTO
    notif_provider = memento.ConsoleNotificationProvider()
    results = memento.Memento(exp_func, notif_provider).run(config_matrix)
\end{lstlisting}
and relax while it runs \texttt{exp\_func} for each task in parallel based on the number of threads available on the computer, caching results, and creating checkpoints. 
The notification provider specifies the notification sent to the user once \memento completes the tasks.

\section{Conclusion}

In this paper, we introduced \memento, a Python library that helps researchers and data scientists run and manage computationally expensive experiments efficiently and reliably. 
\memento's simple configuration matrix and parallelization capabilities allow for running large numbers of experiments quickly. 
The ability to checkpoint in-progress experiments adds convenience and reliability to the process, making it easier to manage long-running experiments. 
While \memento is particularly well-suited to running machine learning experiments, it can be used for any type of laborious and time-consuming experimental pipeline. 




{
    \bibliographystyle{splncs04}
    \bibliography{memento.bib}
}

\end{document}